\documentclass[review, 3p]{elsarticle}  % Comment this line out if you need a4pape

\usepackage{graphics} % for pdf, bitmapped graphics files
\usepackage{graphicx}

\usepackage{color}

\usepackage{amsmath}
\usepackage{algorithm}
\usepackage{algpseudocode}
\usepackage{amsfonts}
\usepackage[utf8]{inputenc}
\usepackage{graphicx} % trim
\usepackage{subfig}
\usepackage{multirow}
\usepackage{footnote}

\usepackage{url}

\usepackage{tabularx, booktabs}
\newcolumntype{Y}{>{\centering\arraybackslash}X}

\usepackage[dvipsnames]{xcolor}

\title{\LARGE \bf
Neural Policy Style Transfer
}

\usepackage[bookmarks=false]{hyperref} %<--- Load after everything else
\author{Raul Fernandez-Fernandez$^*$
}

\author{Juan G. Victores %
}

\author{Jennifer J. Gago}

\author{David Estevez}

\author{\\ Carlos Balaguer%
}

\address{All of the authors are members of the Robotics Lab research group within the Department \\ of Systems Engineering and Automation, Universidad Carlos III de Madrid (UC3M), \\ Madrid, Leganes, Av. Universidad 30, 28911, Spain.}

\cortext[cor0]{Corresponding author: rauferna@ing.uc3m.es}

\begin{document}

%%%%%%%%%%%%%%%%%%%%%%%%%%%%%%%%%%%%%%%%%%%%%%%%%%%%%%%%%%%%%%%%%%%%%%%%%%%%%%%%
\begin{abstract}

Style Transfer has been proposed in a number of fields: fine arts, natural language processing, and fixed trajectories. We scale this concept up to control policies within a Deep Reinforcement Learning infrastructure. Each network is trained to maximize the expected reward, which typically encodes the goal of an action, and can be described as the \emph{content}. The expressive power of deep neural networks enables encoding a secondary task, which can be described as the \emph{style}. The Neural Policy Style Transfer (NPST)\footnote{NPST: Neural Policy Style Transfer} algorithm is proposed to transfer the \emph{style} of one policy to another, while maintaining the \emph{content} of the latter. Different policies are defined via Deep Q-Network architectures. These models are trained using demonstrations through Inverse Reinforcement Learning. Two different sets of user demonstrations are performed, one for \emph{content} and other for \emph{style}. Different \emph{styles} are encoded as defined by user demonstrations. The generated policy is the result of feeding a \emph{content} policy and a \emph{style} policy to the NPST algorithm. Experiments are performed in a catch-ball game inspired by the Deep Reinforcement Learning classical Atari games; and a real-world painting scenario with a full\nobreakdash-\hspace{0pt}sized humanoid robot, based on previous works of the authors. The implementation of three different Q-Network architectures (Shallow, Deep and Deep Recurrent Q-Network) to encode the policies within the NPST framework is proposed and the results obtained in the experiments with each of these architectures compared. 
\end{abstract}

%%%%%%%%%%%%%%%%%%%%%%%%%%%%%%%%%%%%%%%%%%%%%%%%%%%%%%%%%%%%%%%%%%%%%%%%%%%%%%%%


\maketitle

%%%%%%%%%%%%%%%%%%%%%%%%%%%%%%%%%%%%%%%%%%%%%%%%%%%%%%%%%%%%%%%%%%%%%%%%%%%%%%%

\section{Introduction}

%Style Transfer aims to transform a certain input, adapting it via a certain \emph{style} without changing the original \emph{content}

The concept behind Style Transfer is to transform the \emph{style} of a certain input without changing it's original \emph{content}.
This \emph{content} is often referred as the  \emph{what}, whereas the \emph{style} as the \emph{how}.
Style Transfer has been applied essentially to three main fields: computer vision with fine arts, relating objects and shapes with painting techniques;
natural language processing, relating the meaning of a text with the specific selection of words; and, finally, fixed trajectories for animation, relating the final motion with the manner or emotion with which the trajectory is performed.

Early works in computer graphics, specifically in the area of creating trajectories for animating figures, apply ``moods'' or ``emotions'' (\emph{style}) to ``base motion'' actions (\emph{content}) by a weighted addition in the frequency-phase domain \cite{Unuma1991}. This method generates fixed trajectories, which are a kind of actions, but do not involve feedback with respect to the state of an agent. Additionally, it relies on periodicity, which in the human domain limits actions to locomotion tasks. Further literature includes the incorporation of signal processing techniques \cite{Amaya1996}, and Hidden Markov Model representations as well as statistical modelling \cite{Brand2000}.

The first works to coin the terms \emph{style} and \emph{content} are in the context of computer vision, concretely in the area of optical character recognition. An explicit bilinear model was used to separate the \emph{style}, which corresponds to the used font or calligraphy, while the \emph{content} is given by the actual letters or graphemes \cite{Tenenbaum1997}. Further works include multilinear modelling techniques, which employ the N-mode singular value decomposition (SVD) tensor extension of the conventional matrix SVD to transform collections of images into spaces where the same bilinear model can be applied.

\begin{figure}[t]
  \centering
  \includegraphics[width=0.65 \textwidth]{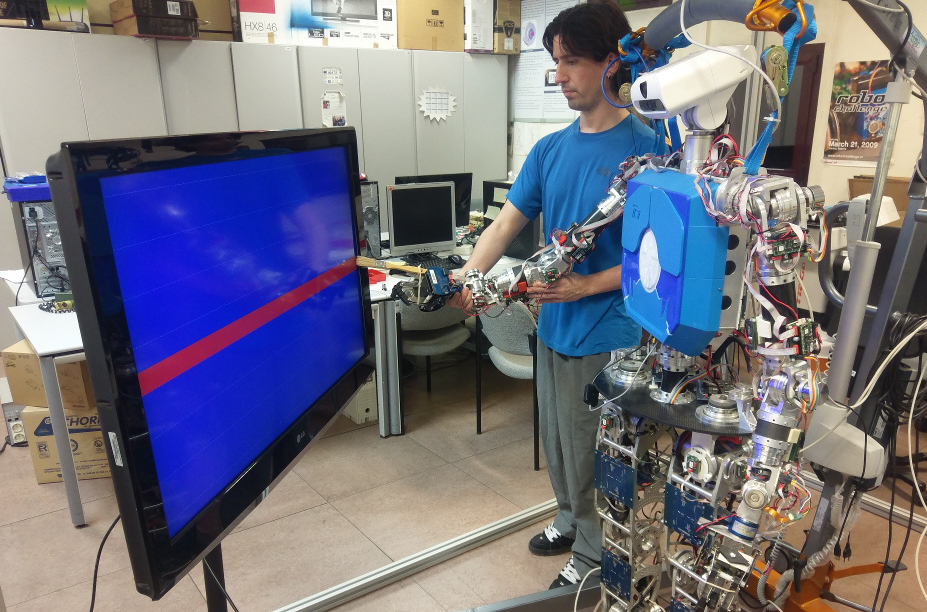}
  \caption{Neural Policy Style Transfer training: ``Grid-world paint'' scenario with TEO the humanoid robot.}
   \label{fig:painting}
\end{figure}

The model known as AlexNet \cite{Krizhevsky2012} demonstrated the potential of Deep Neural Networks (DNN) for computer vision image  classification tasks, and a large variety of alternative DNN models spawned:

On the one hand, VGG \cite{Simonyan2014} became particularly popular for image classification. The VGG-19 network, pretrained on the ImageNet dataset, was used by Gatys et al~\cite{Gatys2016} to develop Style Transfer for images, which marked the beginning of its application to fine arts. In this case, Style Transfer is developed as an optimization process where the pretrained weights of the VGG-19 network remain constant throughout all the iterations. Three images are used as the input of the network: These are the \emph{content} image, the \emph{style} image, and the generated image.
The generated image is the result of optimizing a loss function consisting on a weighted sum of \emph{content} loss and \emph{style} loss. 
The \emph{content} loss is obtained as the difference between the outputs of the high level features of the network corresponding to the \emph{content} and the generated image. The \emph{style} loss is defined using the ``Gram matrix'' of the \emph{style} and generated image, which is based on monitoring the activation values across channels and through different layers. Following the ideas proposed by Gatys et al, several works have been proposed in different areas. Dumoulin et al. \cite{Dumoulin2016} proposed the introduction of a parametric generalizable DNN for Style Transfer in images able to encode multiple styles. Fu et al. \cite{Fu2017} proposed two different algorithms using Autoencoders to introduce Style Transfer in natural language processing without using parallel data. In Lee et al. \cite{Lee2019}, a Style Transfer method encoding the \emph{style} as a noise introduced in the text is proposed. Two neural models are trained, one to introduce the style and other to remove the style and generate the clean text. Other applications include: the introduction and merging of new elements in artistic images \cite{Luan2018}; portrait Style Transfer using soft masks \cite{zhao2020portrait} \cite{zhao2020automatic}; and fixed trajectory generation for animation as proposed by \cite{Holden2017}.

On the other hand, Deep Reinforcement Learning (DRL)~\cite{Mnih2013} also arose as part of the trend, applying DNN to Reinforcement Learning, where agents learn control policies to maximize perceived rewards. Initial applications were video-game oriented, and the rewards were equivalent to the obtained scores. Temporal aspects of the actions were taken into consideration using as input a concatenation of $k$ images corresponding to the last $k$ time steps. DRL marks the beginning of DNN applied to control policies, using the network to represent the Q-value function of the Reinforcement Learning problem statement. Modern DRL approaches include Deep Q-Networks (DQN), Trust Region Policy Optimization, Generative Adversarial Imitation Learning, and Asynchronous Advantage Actor-Critic \cite{Arulkumaran2017}.

Drawing inspiration from the two main previously stated sources, emerged from the outbreak of DNN, this paper presents a Neural Policy Style Transfer (NPST) algorithm. NPST is proposed with the goal to improve the generalization capabilities of robots with the introduction of \emph{styles} as a way to achieve action adaptation. The same base action can be adapted to be used in different contexts with the introduction of different recorded \emph{styles}. In addition to this, a wide range of applications can be derived from the contributions proposed by this paper and summarized in the following three points:

\begin{itemize}
    \item The introduction of Inverse Reinforcement Learning (IRL) algorithms to encode the \emph{content} and \emph{style} using DQN as defined by human demonstrations.        
    \item Proposal of the Neural Policy Style Transfer (NPST) algorithm to perform Style Transfer between policies, applicable to Q\nobreakdash-\hspace{0pt}value functions expressed as DQN.
    \item Two different experimental scenarios: the ``Catch-ball'' scenario inspired by the Deep Reinforcement Learning classical Atari games, where a ball must be caught by a side-by-side moving paddle; and the ``Grid-world paint'' scenario, a painting scenario with a full-sized humanoid robot, which builds upon previous experiments in evolutionary cognitive robotics \cite{Fernandez-Fernandez2018} \cite{fernandezfernandez2018robot}  and is equivalent to a grid-world problem in the real world.
\end{itemize}

The remainder of this paper is organized as followed: Section \ref{BG} provides a Reinforcement Learning background. Section \ref{ST} describes the proposed Style Transfer algorithm. Sections \ref{sec:experiments}, \ref{experiments1}, and \ref{experiments2} depict the experiments performed in this paper. Finally, conclusions are drawn in Sections \ref{sect:Discussion} and \ref{conclusions}.

\section{Background}
\label{BG}

Let a Markov Decision Process (MDP) be defined using the tuple $M = \{S,A,T,\gamma, D, R \}$, where $S$ represents the state space, $A$ the action space, $T=\{P_{sa}\}$ the transition probabilities defined by the environment, where $P_{sa}$ is the state transition distribution upon taking action $a$ in state~$s$, $\gamma \in [0,1]$ is the discount factor, $D$ is the initial state distribution, and $R : S \rightarrow{} A$ the reward function. An optimal policy $\pi^*(s)$ can be found such as the reward obtained by the agent is maximized over a full execution of the problem. This policy defines the behavior of the agent mapping actions with states. A Q-learning approach \cite{Watkins1992} defines this policy $\pi^*(s)$ using the Q-value function defined in Eq. \ref{eq:Q-function}. 

\begin{equation}\label{eq:Q-function}
\begin{split}
Q(S_{t},A_{t}) = Q(S_{t},A_{t}) + \alpha[R_{t+1}+\gamma \max_{a}Q(S_{t+1},A)-Q(S_{t},A_{t})]
\end{split}
\end{equation}

where $t$ can be any time step, $\alpha$ is the learning rate, and $Q(S_{t},A_{t})$ defines the expected reward obtained over a full episode starting from the state $s$ with an action $a$ and following a greedy policy. In DQN, this Q-function is encoded using a DNN.

The definition of a proper reward function is a critical step of Reinforcement Learning architectures that defines the behavior of the obtained agent. Inverse Reinforcement Learning (IRL) is proposed as a way to define the reward function $R$ using a set of $m$ expert demonstrations $E = \{s_0^{(i)},s_1^{(i)},...\}_{i=1}^m$. 

The IRL algorithm, as defined by Abbeel et al \cite{abbelng2004}, assumes that a $k$-dimensional feature vector $\phi(s) \in [0,1]$ exists in $S$ such that $R$
can be defined as
$R(s) = w \cdot \phi(s)$,
where $w \in \mathbb{R}^k[0,1] $ is defined as the weight feature vector to optimize. While the feature vector $\phi(s)$ can be hand-crafted by the designer, some approaches have been proposed for its selection to be automated \cite{Wulfmeier2015}. 

The expert feature expectation $\hat{\mu}_E$ can be defined as in Eq.~\ref{expected_features_expert}:

\begin{equation}\label{expected_features_expert} 
\hat{\mu}_E = \dfrac{1}{m}\sum_{i=1}^{m}\sum_{t=0}^{\infty}\gamma^t\phi(s_t^{(i)})
\end{equation}

This expert feature expectation $\hat{\mu}_E$ depicts a measure of the degree of desirability of the different features for the demonstrator. This provides an intuition of which features are related to larger and lower rewards and, therefore, allows the estimation of the reward as a function of these features. 

Given an MDP where the reward function
$R(s)$
is unknown, the objective of the IRL algorithm is to find an optimal policy $\pi^*$, defined by $R(s)$, such that it satisfies Eq.~\ref{irl_eq}:
\begin{equation}\label{irl_eq} 
||\mu(\pi^*)-\hat{\mu}_E ||_{2} < \epsilon
\end{equation}

where $\mu(\pi^*)$ is defined as in Eq.~\ref{mu_pi}:

\begin{equation}\label{mu_pi} 
\mu(\pi^*) = \mathbb{E}\Bigg[\sum_{t=0}^{\infty}\gamma^t\phi(s_t)\Bigg|\pi^*\Bigg] \in \mathbb{R}^k
\end{equation}

Based on these premises, several approaches have been proposed to solve the IRL problem. In this paper, we use the Maximum Entropy IRL approach \cite{ziebart2008maximum}, where the IRL algorithm is reduced to the maximization of a likelihood distribution defined as in Eq.~\ref{MEF}:

\begin{equation}\label{MEF} 
\mathcal{L}(w) = logP(E,w|R) = \mathcal{L}_E+\mathcal{L}_w
\end{equation}

where $\mathcal{L}_E$ and $\mathcal{L}_w$ are, respectively:

\begin{equation}\label{MEF2} 
\mathcal{L}_E=logP(E|R)
\end{equation}
\begin{equation}\label{MEF3}
\mathcal{L}_w=logP(w)
\end{equation}

Wulfmeier et al \cite{Wulfmeier2015} adapted the distribution defined by Eq.~\ref{MEF} to work with DNN by defining the gradient of the reward function with respect to the weights obtained using backpropagation as in the Eq.~\ref{DMEF1} extracted from Eq.~\ref{DMEF0}:

\begin{equation}\label{DMEF0}
\dfrac{\delta\mathcal{L}}{\delta w} = \dfrac{\delta\mathcal{L}_E}{\delta w} + \dfrac{\delta\mathcal{L}_w}{\delta w}
\end{equation}

where $\dfrac{\delta\mathcal{L}_E}{\delta w}$ 
is given by Eq.~\ref{DMEF1}:

\begin{equation}\label{DMEF1}
\dfrac{\delta\mathcal{L}_E}{\delta w} = \dfrac{\delta\mathcal{L}_E}{\delta R} \cdot \dfrac{\delta R}{\delta w} 
= (\hat{\mu}_E - \dfrac{1}{m}\sum_{i=1}^{m}\sum_{t=0}^{\infty} P(s_t^{(i)} | R) ) \cdot \dfrac{\delta R}{\delta w}
\end{equation}

The gradient of the expert demonstration term $\mathcal{L}_E$ with respect to the model parameters of a linear function is equal to the feature expectation difference along the expert trajectories \cite{ziebart2008maximum}. 
The used DNN model and the Style Transfer algorithm is described in the following section.

\section{Neural Policy Style Transfer}
\label{ST}

Let $\mathcal{C}$ and $\mathcal{S}$ be two different DQN encoding two different actions defined via user demonstrations, which in turn define their control policies $\pi_c$ and $\pi_s$.
The input must correspond to the observation or state space, and the output must express the Q-value function.

\begin{figure}[htpb]
  \centering
  \includegraphics[width=0.65 \textwidth]{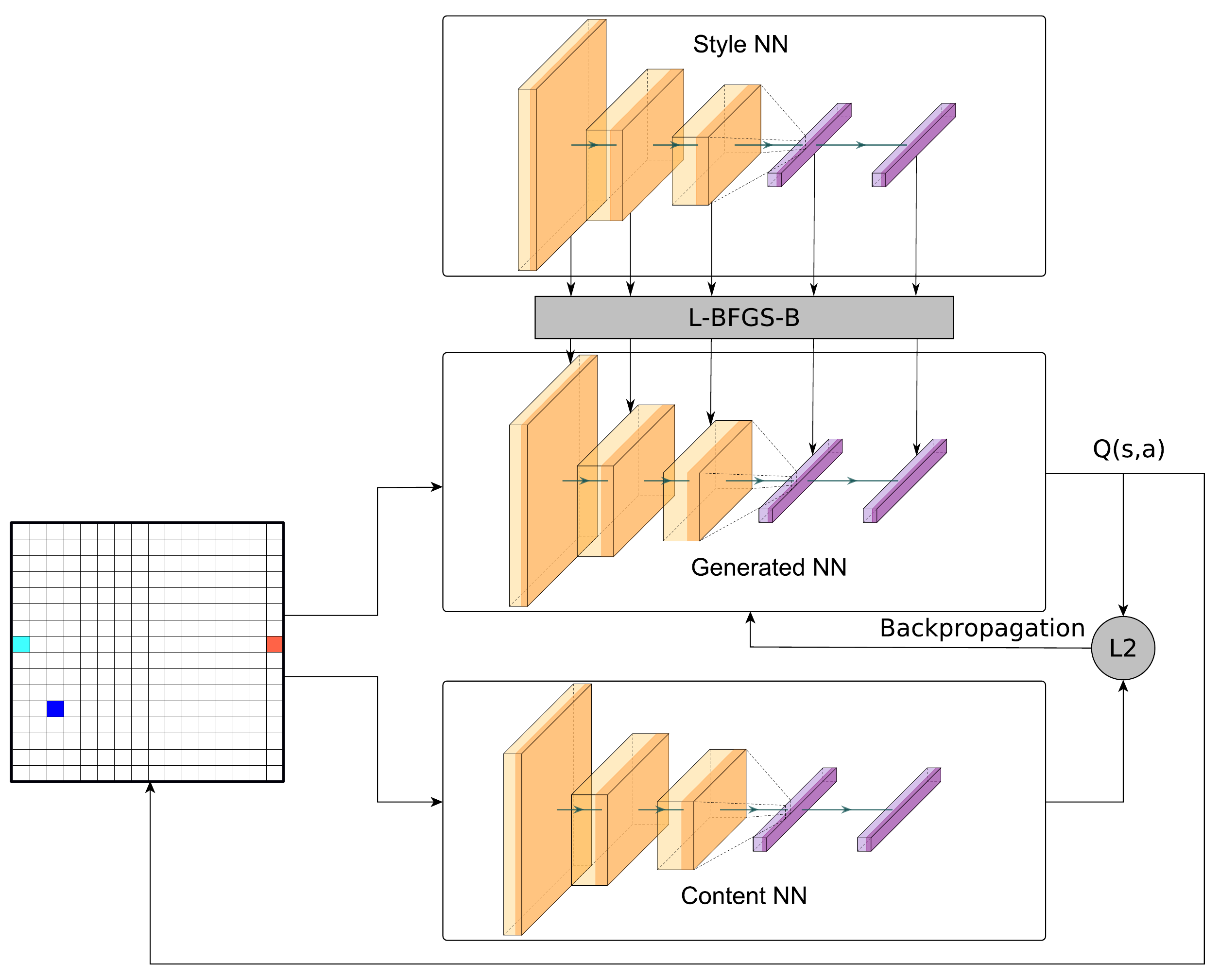}
  \caption{Neural Policy Style Transfer framework. The framework is composed by three Deep Q-Networks with the same architecture. The content DQN is trained using the content demonstrations. The style DQN is trained using the style demonstrations. The Generated DQN is generated using the output of the Content DQN and the weights of the Style DQN.}
   \label{fig:model}
\end{figure}

Let $\mathcal{G}$ be the DQN that encodes the generated action, the output of the Neural Policy Style Transfer (NPST) algorithm, which in turn defines the control policy $\pi_g$. $\mathcal{G}$ can be defined as a combination of the \emph{content} and \emph{style} defined by the two base DQN, $\mathcal{C}$ and $\mathcal{S}$. 
%The \emph{content} can be extracted from the action $C$ while the \emph{style} can be extracted from the action $S$. 
The \emph{content} can be defined as the high level features of $C$ encoding the goal of the action. The \emph{style} can be defined as the weights of $S$ defining the ``mood'' or ``emotion'' as defined by user demonstrations.
%$\mathcal{C}$ can be defined as an action that determines the goal (\emph{content}) of $\mathcal{G}$, whereas $\mathcal{S}$  determines the ``mood'' or ``emotion'' secondary task (\emph{style}) in which the action $\mathcal{G}$ will be performed. 
%The NPST algorithm is proposed to transfer the style of $\mathcal{S}$ and the content of $\mathcal{C}$ to a new network $\mathcal{G}$.

An approach analogous to the one proposed by Gatys et al~\cite{Gatys2016}
would involve optimizing to minimise a weighted sum of a \emph{content} loss
and a \emph{style} loss. However, there are significant differences in the meaning of each element involved.
Caution must be taken so at least one of these losses is clearly separated from the other, to avoid straying into a simple \emph{mix} of actions.

The \emph{content} transfer is performed by training $\mathcal{G}$ based on the high-level features of $\mathcal{C}$, corresponding to the Q-value output.
A backpropagation step is performed over $\mathcal{G}$ using the output of $\mathcal{C}$ as the true label. A Mean Squared Error (MSE) function is used as the loss function for the backpropagation algorithm. This is equivalent to defining the \emph{content} loss $\mathcal{L}_{content}$ for a single output as in Eq. \ref{c_loss}:

\begin{equation}\label{c_loss}
\mathcal{L}_{content}(\mathcal{G},\mathcal{C}) = ||q_{\mathcal{G}}-q_{\mathcal{C}}||_2
\end{equation}

Where $q_{\mathcal{G}}$ and $q_{\mathcal{C}}$ are the high level features of $\mathcal{G}$ and $\mathcal{C}$. These are the DQN outputs of $\mathcal{G}$ and $\mathcal{C}$ for a given state.

Its counterpart, the \emph{style} transfer of the algorithm is implemented via an optimization that depends on the \emph{style} loss $\mathcal{L}_{style}$, which is defined as in Eq. \ref{s_loss}:

\begin{equation}\label{s_loss}
 \mathcal{L}_{style}(\mathcal{G},\mathcal{S}) = ||w_{\mathcal{G}}-w_{\mathcal{S}}||_2
\end{equation}

where $w_\mathcal{G}$ and $w_\mathcal{S}$ correspond to the weights of the models that represent $\mathcal{G}$ and $\mathcal{S}$, respectively. 

The assumption here is that by introducing the \emph{style} transfer through the weights of the network, the framework is giving preference to the outputs preferred by the \emph{content}. The \emph{style} is introduced as a secondary task in the \emph{content} execution.

\begin{algorithm} [h]
\caption{Neural Policy Style Transfer (NPST)}\label{st_alg}
\begin{algorithmic}[1]
\item[]
\Procedure{NPST}{$\mathcal{C}, \mathcal{S}$, $env$, N}
\item[]
\State \textbf{Initialise:}
\State $\mathcal{G} \gets \mathcal{S}$
\State $w_{\mathcal{S}} \gets \mathcal{S}.get\_weights()$
\State $env.init()$
\State $state\gets env.observe()$
\item[]
\For{n=1:N}
\State \textbf{Update Environment:}
\State $q_\mathcal{G} \gets \mathcal{G}.predict(state)$
\State $a_\mathcal{G} \gets argmax_a (q_\mathcal{G})$
\State $env.step(a_\mathcal{G})$
\State $state\gets env.observe()$
\item[]
\State \textbf{Content Transfer:}
\State $q_\mathcal{C} \gets \mathcal{C}.predict(state)$
\State $q_\mathcal{G} \gets \mathcal{G}.predict(state)$
\State $\mathcal{G}.backprop(q_\mathcal{G}, q_\mathcal{C})$
\item[]
\State \textbf{Style Transfer:}
\State $w_{\mathcal{G}} \gets \mathcal{G}.get\_weights()$
\State $w_{\mathcal{G}} \gets \textsc{l-bfgs-b}(||w_{\mathcal{G}}-w_{\mathcal{S}}||_2)$
\State $\mathcal{G}.set\_weights(w_{\mathcal{G}})$
\EndFor
\item[]
\item[]
\EndProcedure
\item[]
\end{algorithmic}
\end{algorithm}

The DQN $\mathcal{G}$ is initialized as a copy of $\mathcal{S}$, both in model as in pre-computed weights.
The environment is initialized and its state is observed. 
$\mathcal{G}$ is 
updated for N iterations:

\begin{enumerate}

\item The environment is updated using the action defined by $\mathcal{G}$.

\item 
The values $q_\mathcal{C}$ and $q_\mathcal{G}$, which represent the high-level features of $\mathcal{C}$ and $\mathcal{G}$ respectively, are obtained for the current state. 
A backpropagation step is then performed over the model of $\mathcal{G}$.

\item 
The weights of $\mathcal{G}$ are updated using a box-constrained limited-memory Broyden–Fletcher–Goldfarb–Shanno (L-BFGS-B) algorithm to minimise the \emph{style} loss as presented in Eq. \ref{s_loss}.
\end{enumerate}

The full algorithm is shown in Alg.~\ref{st_alg}.

\section{Experiments}
\label{sec:experiments}

Three different neural network architectures are introduced in the experiments to measure the performance of the NPST algorithm. The first architecture is a DQN with the same architecture as the one proposed by Mnih et al \cite{Mnih2015a}. The second architecture, referred as Shallow Q-Network (SQN), is a smaller version of the first DQN architecture. In SQN, the second and third Convolutional Layers (CL) of the DQN architecture are removed. The same Fully Connected (FC) layers are used for both architectures. The last architecture introduced in the experiments is a Deep Recurrent Q-Learning Network (DRQN), referred as DRQN, with the same architecture as the one proposed by Brejl et al. \cite{Rasmus2018}. The size of all the architectures is the same and corresponds to the size of the layers defined by Mnih et al. \cite{Mnih2015a}. The Long Short-Term Memory (LSTM) layer defined with the DRQN architecture is composed by 256 nodes.

Two different experimental scenarios are proposed to measure the performance of these three architectures with the NPST algorithm. The ``Catch-ball'' experimental scenario is inspired by classical DRL scenarios using Atari games. The ``Grid-world'' paint scenario is designed to work with a humanoid robot and based on previous works of the authors \cite{Fernandez-Fernandez2018} \cite{fernandezfernandez2018robot}.

\section{``Catch-ball'' experiment}
\label{experiments1}

The first experiment consists in a ``Catch-ball'' game scenario, similar to the Pong arcade game. A ball is released from a random location from the top of the screen, and falls vertically. The agent can move a paddle horizontally at the bottom of the screen. The agent wins if it catches the ball with the paddle before it falls off the screen.

\subsection{Experimental setup}

Three different sets of five expert demonstrations are performed in these experiments. These sets correspond to three different actions: 
one \emph{content} action, and two different \emph{style} actions (``nervous'' and ``fall''). The \emph{content} action aims to win the game by fulfilling the goal of catching the falling ball by using the paddle. Both demonstrated \emph{style} actions ignore the position of the ball. The first \emph{style} imitates a ``nervous'' behaviour or mood, tending to perform small moves around the same position.
The second  \emph{style} imitates a ``fall'' movement, always tending towards the left side of the screen. 

The reward functions that define these actions are obtained using the Maximum Entropy Deep IRL algorithm presented in section \ref{BG}. These reward functions are used to train the networks that are introduced in the NPST algorithm. For the training of the IRL algorithm, a hand-crafted feature vector $\phi(s)$ and latent state space $s$ are selected for the three sets of demonstrations. In the case of the \emph{content} demonstrations, $s$ is defined as a function of ball and paddle positions, and $\phi(s)$ classifies if the paddle and ball positions are aligned. In the case of the \emph{style} demonstrations, the same latent state space $s$ was used for both \emph{styles}. This space is defined using the last three paddle positions, corresponding to the last three time steps of the paddle. 
The feature vector $\phi(s)$ used for the \emph{styles} encodes the spatial-temporal information  of the paddle. In the case of the ``nervous'' \emph{style}, $\phi(s)$ classifies if a movement was performed with the same starting and ending position. In the case of the ``fall'' \emph{style}, $\phi(s)$ classifies if a movement to the left was performed.

For the \emph{content} demonstrations, 5 iterations of the IRL algorithm were performed, while 2 iterations were used for the \emph{style} demonstrations. The IRL discount factor ($\gamma$) used was 0.9 with a learning rate of 0.01 for all the actions. Three reward functions $R(s)$ are obtained corresponding to the three base actions (\emph{content}, ``nervous'' and ``fall''). For each of these base actions, three neural networks, corresponding to the three architectures proposed, are trained using the same $R(s)$. These networks are trained using Q-learning and referred in the experimental results as the Vanilla neural networks. These Vanilla neural networks are the base neural networks that will later be used to define the \emph{content} and {style} in the NPST framework. Experimental results obtained with these Vanilla networks are added as a baseline. The same Vanilla Content neural networks are introduced for the transferring of both of the \emph{styles}. The input of the networks is the raw 80x80 pixel image of the screen, and the outputs correspond to the Q-values assigned to the three possible actions (stay still, go left, and go right).  

\begin{table}[h!]
\vspace{0.5em}
\centering
\caption{Hyperparameters for the ``Catch-ball'' scenario.}
\label{tab:catch-ball}
\begin{tabular}{ll}
%\hline
Hyperparameter                 & Setting                             \\ \hline
\emph{Shared between architectures}                          &                                     \\
\hspace{3ex}Activation                            & ReLU \cite{nair2010}   \\ 
\hspace{3ex}Initialization                             & Normal distribution  \\

\emph{DQN architecture}                          &                                     \\
\hspace{3ex}Layers                            & (CL, CL, CL, FC, FC)   \\
\hspace{3ex}Layers Configuration (Size, Kernel, Strides)                      & ((32,8,4),(64,4,2),(64,3,1),(512,-,-),(3,-,-))    \\
%\hspace{3ex}Kernel                            & [8,4,3]    \\
%\hspace{3ex}Strides                           & [4,2,1]    \\

\emph{SQN architecture}                          &                                     \\
\hspace{3ex}Layers                            & (CL, FC, FC)   \\
\hspace{3ex}Layers Configuration (Size, Kernel, Strides)                       & ((32,8,4),(512,-,-),(3,-,-))    \\
%\hspace{3ex}Kernel                            & [8]    \\
%\hspace{3ex}Strides                           & [4]    \\

\emph{DRQN architecture}                          &                                     \\
\hspace{3ex}Layers                            & (CL, CL, CL, LSTM, FC)   \\
\hspace{3ex}Layers Configuration (Size, Kernel, Strides)                       & ((32,8,4),(64,4,2),(64,3,1),(256,-,-),(3,-,-))    \\
%\hspace{3ex}Kernel                            & [8,4,3]    \\
%\hspace{3ex}Strides                           & [4,2,1]    \\

\emph{Q-learning and NPST}                         &                                     \\
\hspace{3ex} Image size                     & 80x80                               \\
\hspace{3ex} Number of input time steps             & 4 (1 for DRQN)                                  \\
\hspace{3ex} Optimizer                      & Adam \cite{Kingma2014}    \\
\hspace{3ex} Loss function                  & Mean Squared Error                  \\
\hspace{3ex} Number of actions              & 3                                   \\
\hspace{3ex} Discount ($\gamma$)                   & 0.99                                \\
\hspace{3ex} Experience Replay size         & 5000                                \\
\emph{Q-learning}                  &                                     \\
\hspace{3ex} Learning Rate                & 1e-6                                 \\
\hspace{3ex} Initial Epsilon                & 0.1                                 \\
\hspace{3ex} Final Epsilon                  & 1e-5                            \\
\hspace{3ex} Epsilon gradient               & Lineal                              \\
\hspace{3ex} Batch size                     & 32                                  \\
\hspace{3ex} Exploration Epochs             & 100                                 \\
\hspace{3ex} Training Epochs                & 1000                                \\
\emph{NPST}                         &                                     \\
\hspace{3ex} Learning Rate                & 0.01                             \\
\hspace{3ex} Number of iterations (N)                & One full Catch-ball episode \\
\hspace{3ex} Batch size                     & 100                                 \\
\hspace{3ex} L-BFGS-B internal iterations & 1                                  
\end{tabular}%
\end{table}

The hyperparameters used for training the networks and performing the NPST algorithm are depicted in Table \ref{tab:catch-ball}.
The results for the generated NPST actions depicted in the next section are the average of 10 repetitions of the NPST algorithm. 

\subsection{Results}

The results obtained with the NPST algorithm are depicted in Table~\ref{catch_table_nerv} for the case of transferring the ``nervous'' \emph{style} and Table~\ref{catch_table_fall} for the case of transferring the ``fall'' \emph{style}. The $\mathcal{L}_{content}$ and $\mathcal{L}_{style}$ are computed following Eq. \ref{c_loss} and Eq. \ref{s_loss} respectively and are the average of the full NPST execution. Results between different architectures in terms of obtained losses may not be comparable due to differences in the architecture (i.e. the number of nodes). The $Nervous$ $Moves$ parameter measures the number of times the agent changed its direction. This parameter was introduced as a way to measure the dynamic behavior of the agent to differentiate the ``fall'' and ``nervous'' \emph{styles}. Finally, the $Wins(\%)$ parameter measures the percentage of wins obtained by the agent. 

A clear increase in the number of $Nervous$ $Moves$ performed by the agent was measured when transferring the ``nervous'' \emph{style} with respect the Vanilla Content and Vanilla Fall networks. At the same time, the number of wins decreased when introducing the \emph{styles} with respect the Vanilla Content networks but increased with respect the Vanilla Style networks. The resulting generated control policy combines the results obtained by the \emph{content} and \emph{style} policies.

\begin{table}[h!]
\centering
\caption{Experimental results for the ``Catch-ball'' action introducing the ``nervous'' \emph{style}.}
\label{catch_table_nerv}
\begin{tabularx}{0.9\textwidth}{|c *{7}{|Y}|}
	 \hline
	 Actions & $\mathcal{L}_{content}$ & $\mathcal{L}_{style}$  & $Nervous$ $Moves$ & $Wins( \%)$ \\ \hline
	 Vanilla Content DQN  & --- & 32857.59 & 75 & 100 \\ \hline
     Vanilla Content SQN  & --- & 16805.58 & 73 & 100 \\ \hline
	 Vanilla Content DRQN  & --- & 33766.86  & 28 & 30 \\ \hline
	 Vanilla Nervous Style DQN  & 1.51 &  --- & 280 & 20 \\ \hline
	 Vanilla Nervous Style SQN  & 1.70 &  --- & 298 & 20 \\ \hline
	 Vanilla Nervous Style DRQN  & 1.77 &  --- & 45 & 20 \\ \hline
	 NPST Nervous Generated DQN   & 0.17 & 0.97 & 160 & 60 \\ \hline
	 NPST Nervous Generated SQN   & 0.32 & 3.03 & 166 & 40 \\ \hline
	 NPST Nervous Generated DRQN   & 0.17 & 1.61 &  57 & 30 \\ \hline
 
	\end{tabularx}

\end{table}

\begin{table}[h!]
\centering
\caption{Experimental results for the ``Catch-ball'' action introducing the ``fall'' \emph{style}.}
\label{catch_table_fall}
\begin{tabularx}{0.9\textwidth}{|c *{7}{|Y}|}
	 \hline
	 Actions  & $\mathcal{L}_{content}$ & $\mathcal{L}_{style}$ & $Nervous$ $Moves$  & $Wins( \%)$ \\ \hline
	 Vanilla Content DQN  & --- & 16816.59 & 73 & 100 \\ \hline
	 Vanilla Content SQN  & --- & 32837.50 & 82 & 50 \\ \hline
	 Vanilla Content DRQN  & --- & 33753.42  & 28 & 30 \\ \hline
	 Vanilla Fall Style DQN  & 1.21 &  --- & 2 & 30 \\ \hline
	 Vanilla Fall Style SQN  & 2.19 &  --- & 14 & 0 \\ \hline
	 Vanilla Fall Style DRQN  & 1.33 &  --- & 10 & 0 \\ \hline
	 NPST Fall Generated DQN   & 0.20 & 0.94 & 48 & 50 \\ \hline
	 NPST Fall Generated SQN   & 0.34 & 3.81 & 63 & 50 \\ \hline
	 NPST Fall Generated DRQN   & 0.16 & 2.54 &  44 & 20 \\ \hline

	\end{tabularx}

\end{table}

\begin{figure*}[h!]
  \centering
  \includegraphics[width=0.64\textwidth]{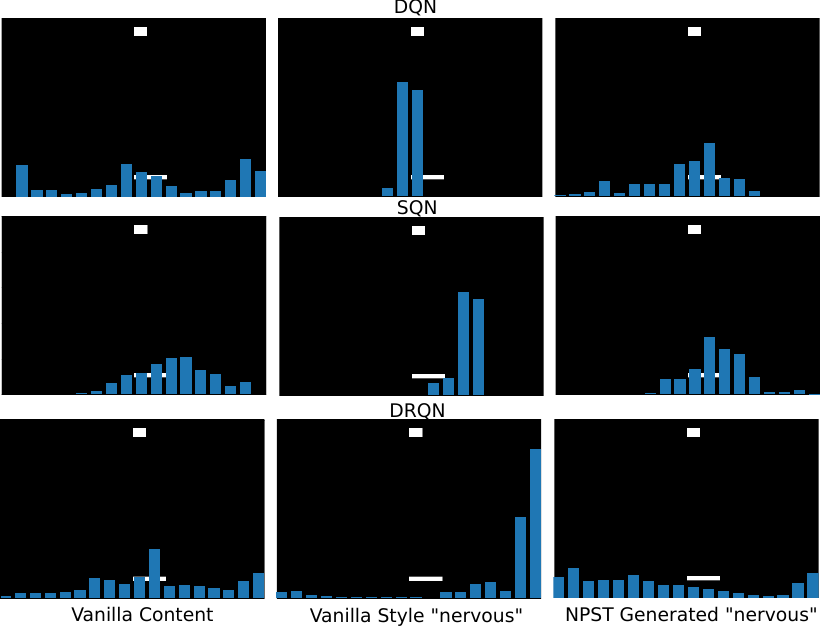}
  \caption{Paddle positions histograms for the case of transferring the ``nervous'' \emph{style}. Each row corresponds to a different network architecture. Each column corresponds to a different action. The Y-axis depicts the number of times each position is visited. All Y-axis are scaled in the range [0, 250]. The X-axis depicts the possible paddle positions within the game environment. }
  \label{fig:catch-histogram-nerv}
  
\end{figure*} 
\begin{figure*}[h!]
  \centering
  \includegraphics[width=0.64\textwidth]{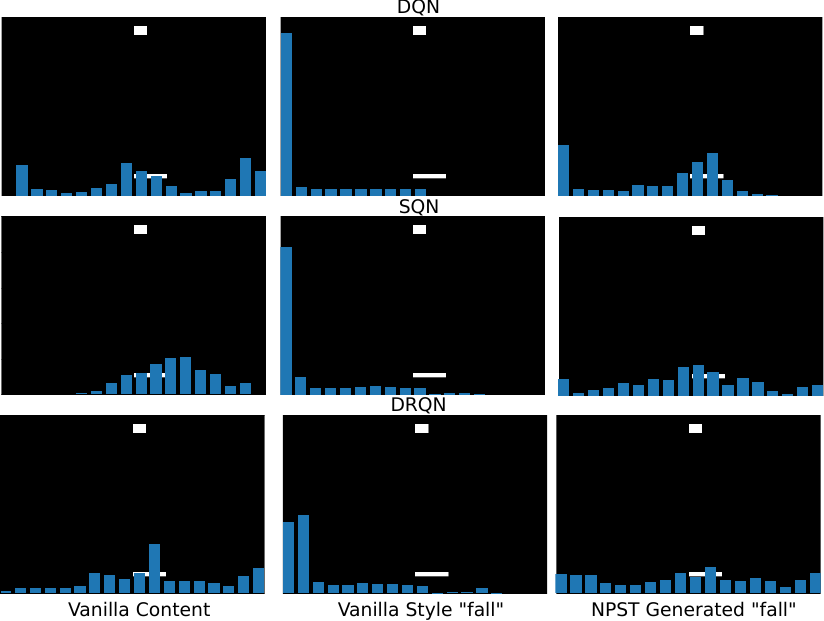}
  \caption{Paddle positions histograms for the case of transferring the ``fall'' \emph{style}. Each row corresponds to a different network architecture. Each column corresponds to a different action. The Y-axis depicts the number of times each position is visited. All Y-axis are scaled in the range [0, 250]. The X-axis depicts the possible paddle positions within the game environment.}
  \label{fig:catch-histogram-fall}
\end{figure*} 

Fig. \ref{fig:catch-histogram-nerv} and Fig. \ref{fig:catch-histogram-fall} depict the number of times each position was visited by the paddle. The depicted results are the cumulative results of the ten repetitions. These figures depict the preferred states of the agent during the experiments. The background image of the game is introduced as reference. The ``nervous'' \emph{style} transferred policies show a preference for the positions in the center. The ``fall'' \emph{style} transferred policies show a preference for the positions in the borders. This tendency is clear in the case of the DQN architecture where the most visited position is at the left of the screen. This corresponds with the behavior shown by the Vanilla Fall networks.

\begin{figure*}[h!]
  \centering
  \includegraphics[width=0.22\textwidth]{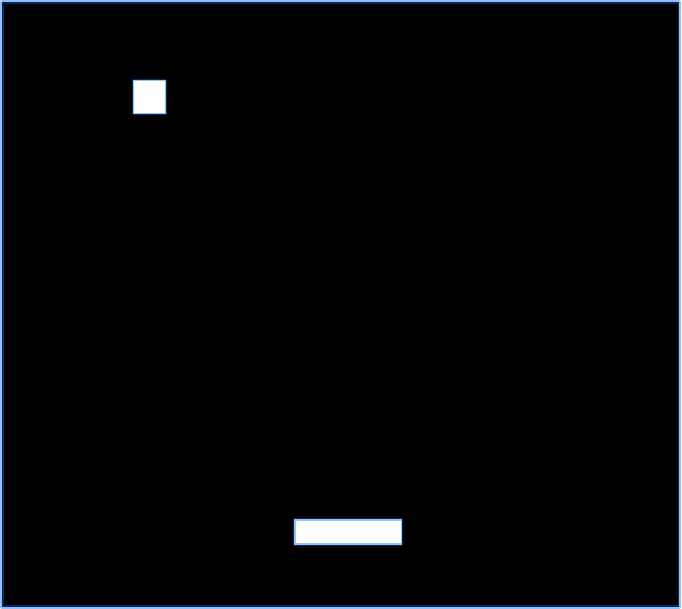}
  \includegraphics[width=0.22\textwidth]{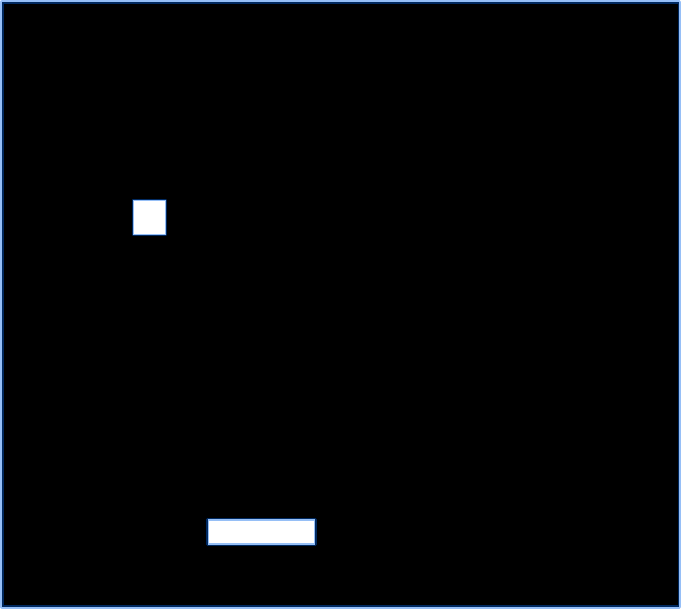}
  \includegraphics[width=0.22\textwidth]{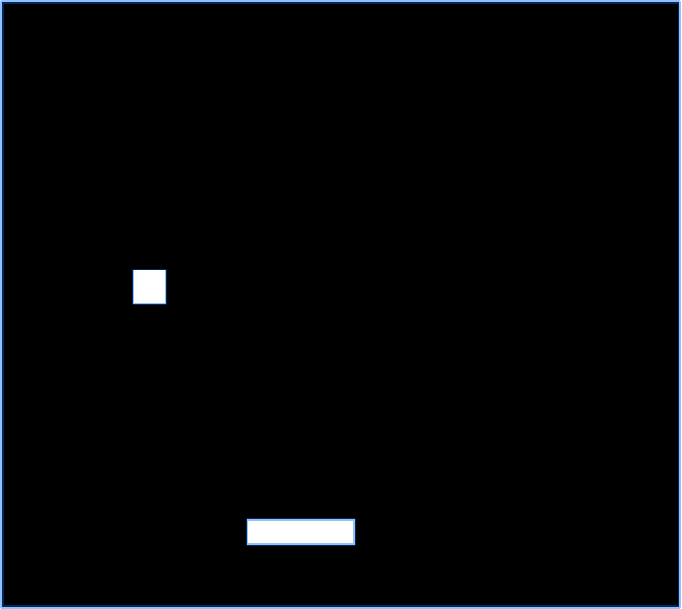}
  \includegraphics[width=0.22\textwidth]{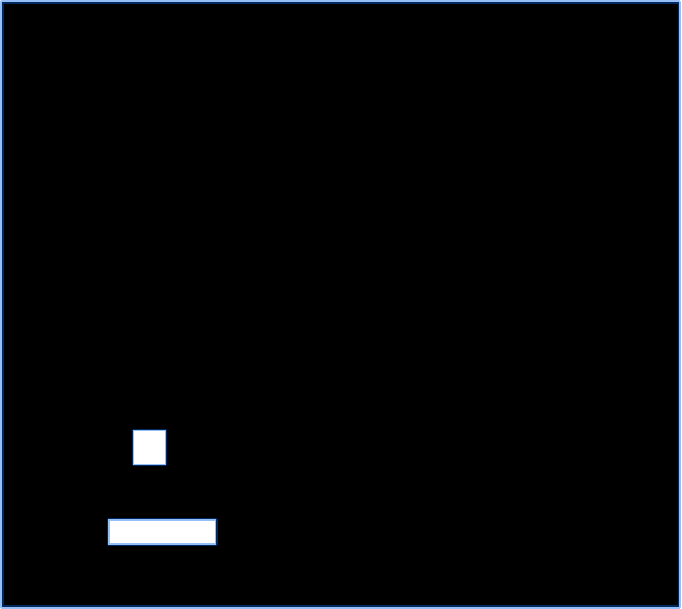} \\
  %\vspace{0.01cm}
  \includegraphics[width=0.22\textwidth]{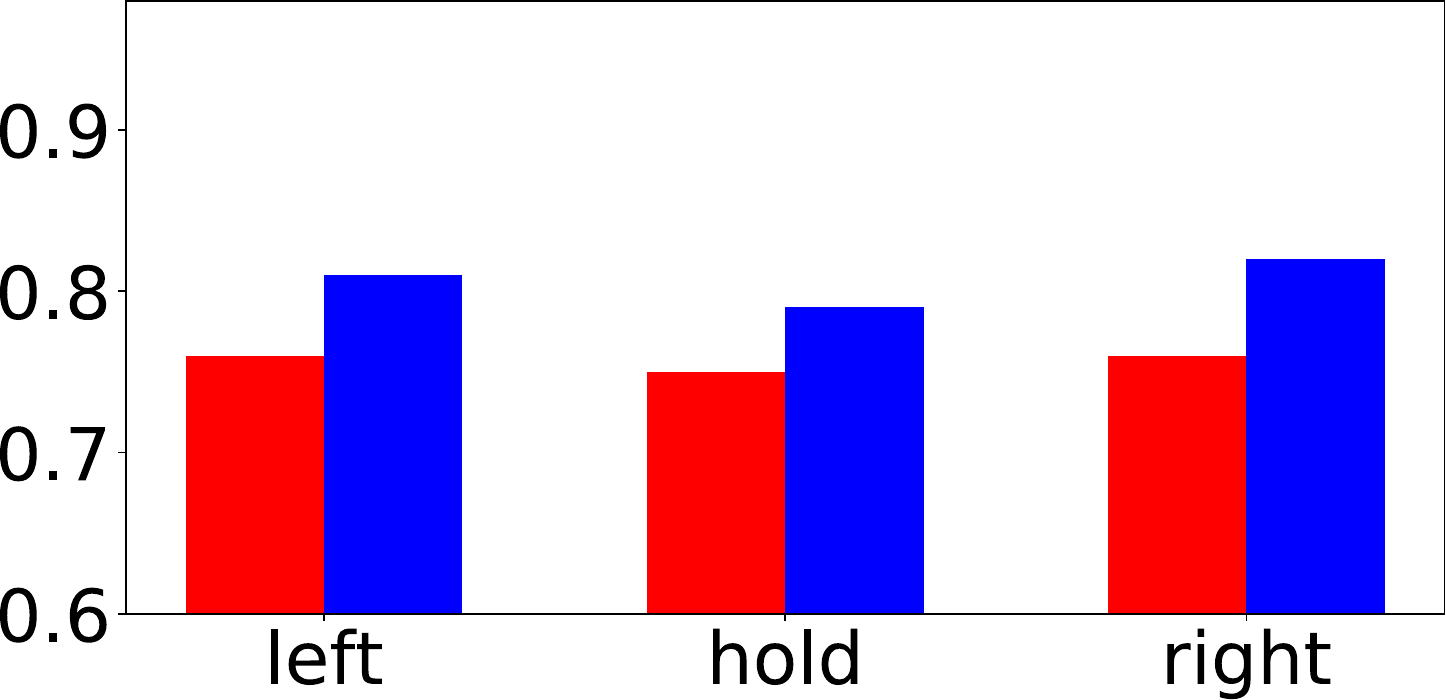}
  \includegraphics[width=0.22\textwidth]{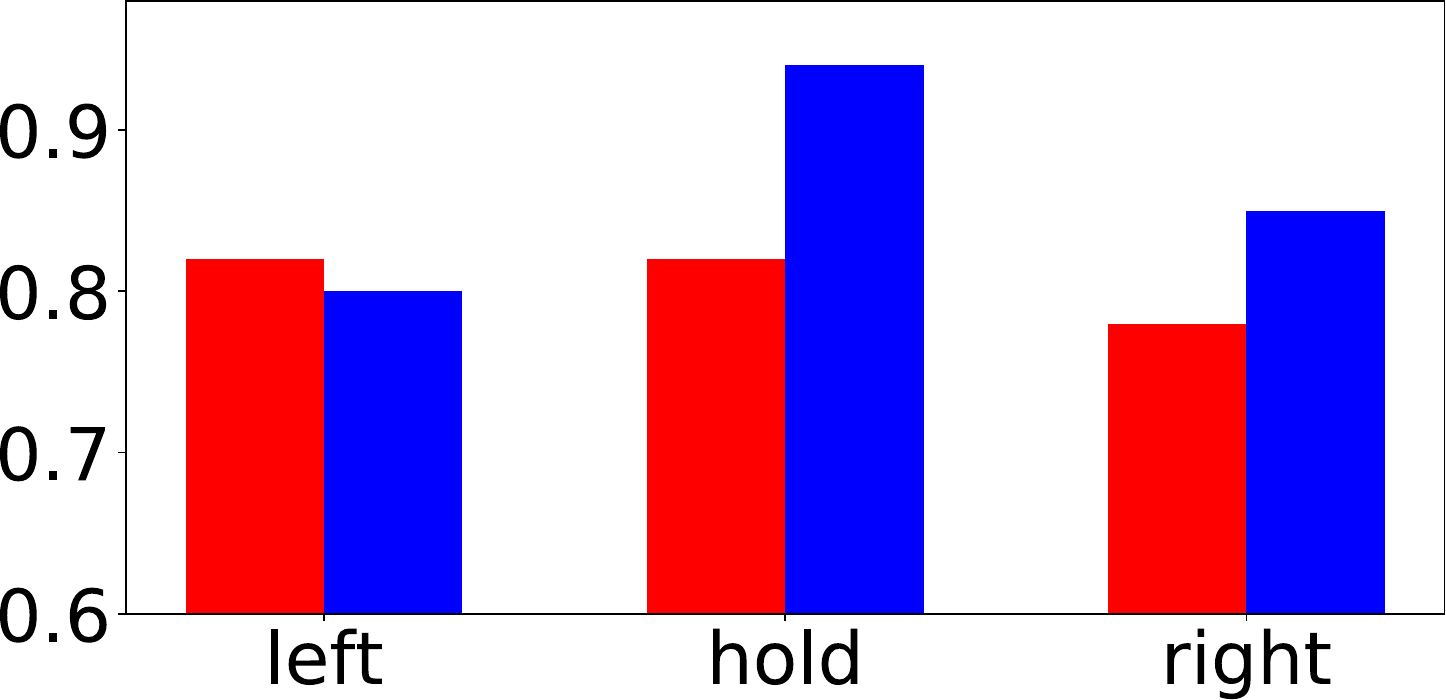} 
  \includegraphics[width=0.22\textwidth]{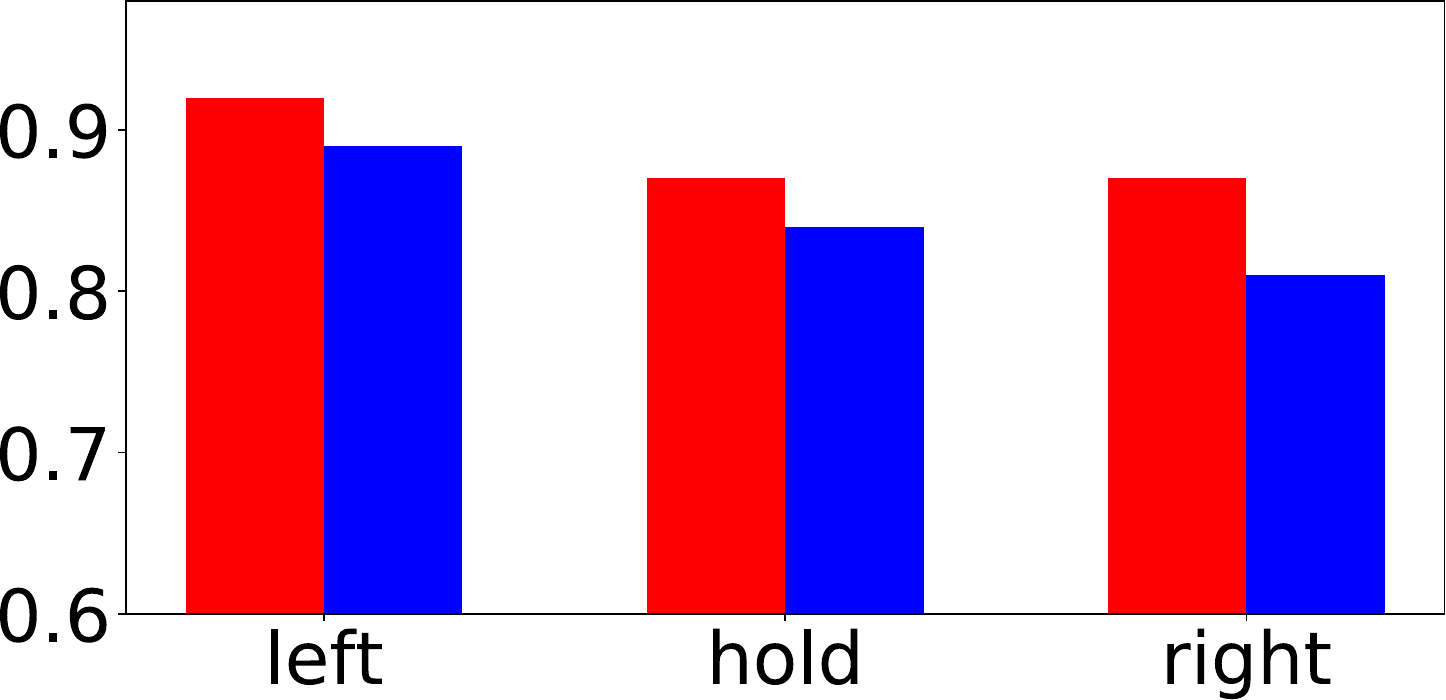}
  \includegraphics[width=0.22\textwidth]{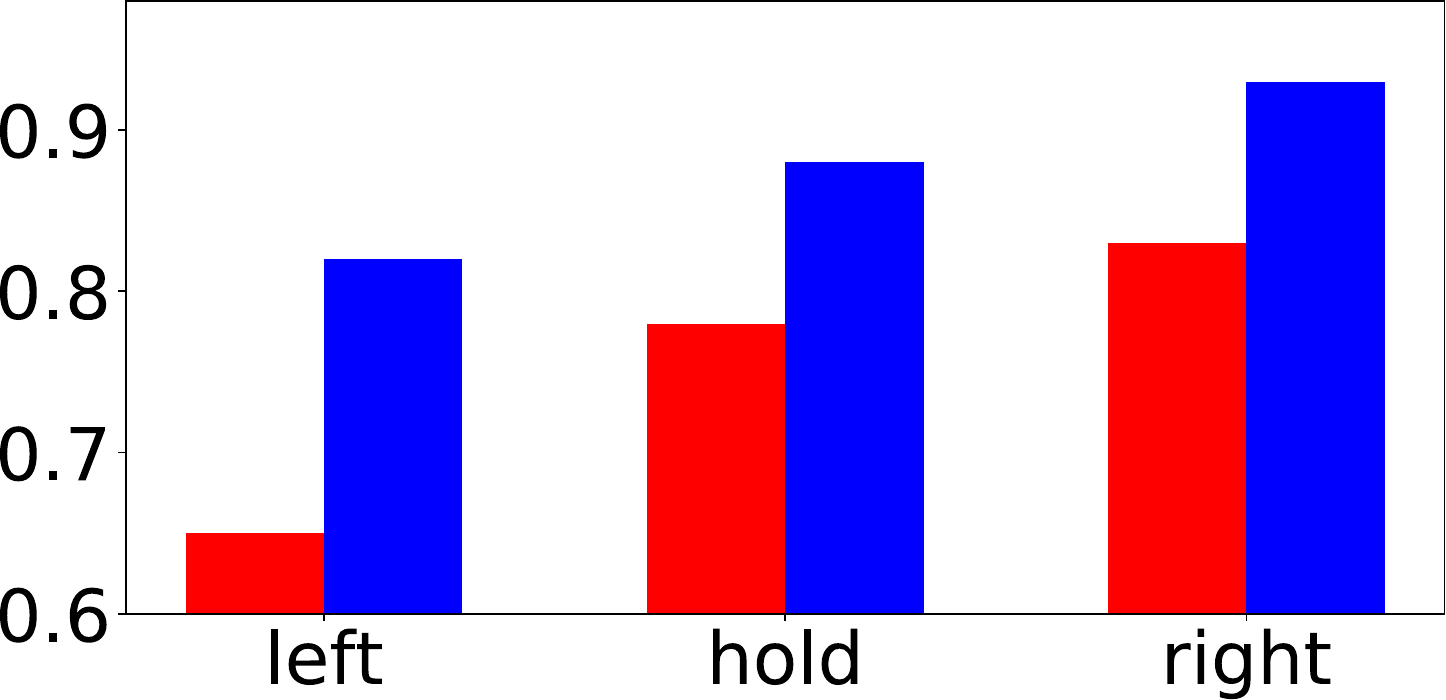} \\
  %\vspace{0.01cm}
  \includegraphics[width=0.22\textwidth]{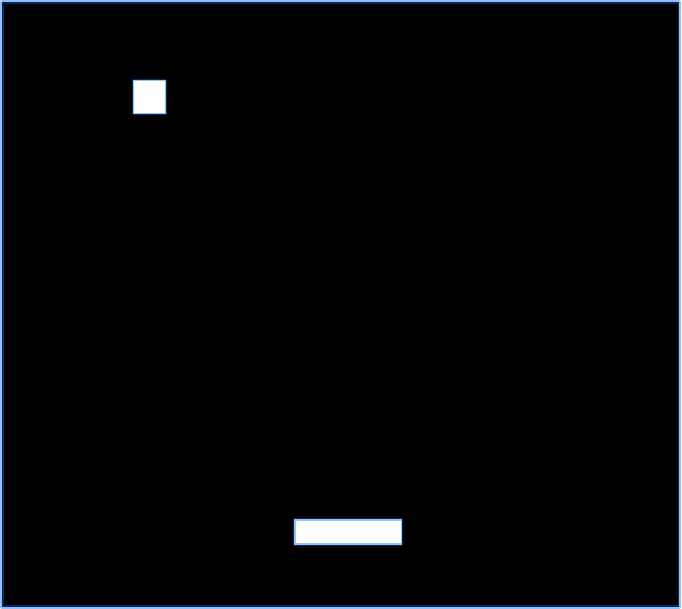}
  \includegraphics[width=0.22\textwidth]{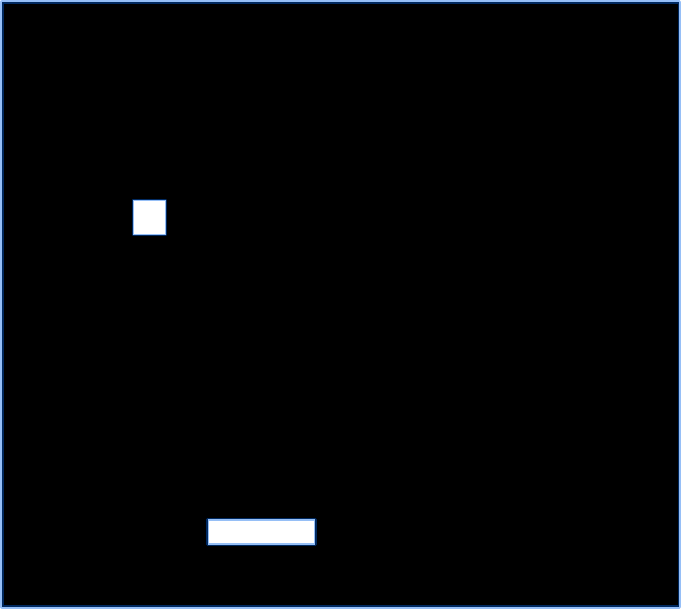}
  \includegraphics[width=0.22\textwidth]{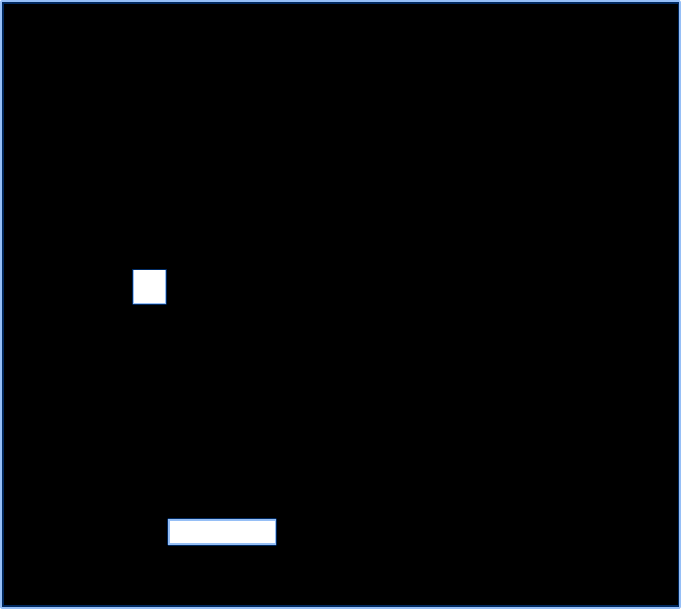}
  \includegraphics[width=0.22\textwidth]{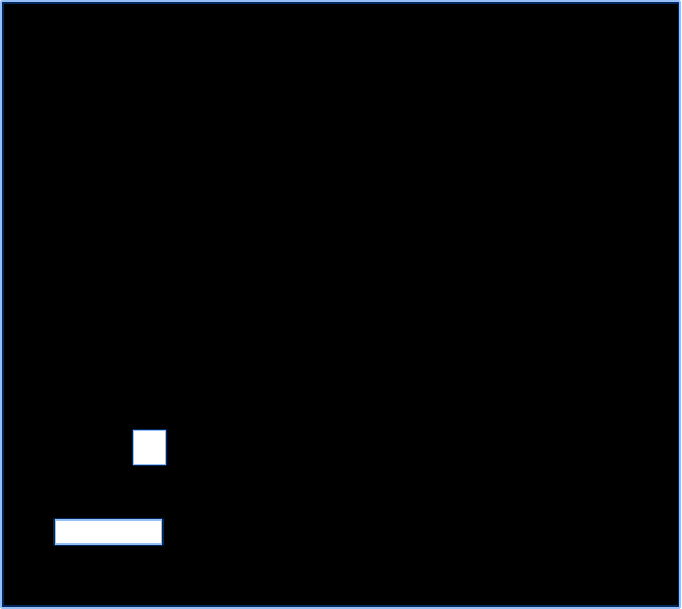} \\
  %\vspace{0.01cm}
  \includegraphics[width=0.22\textwidth]{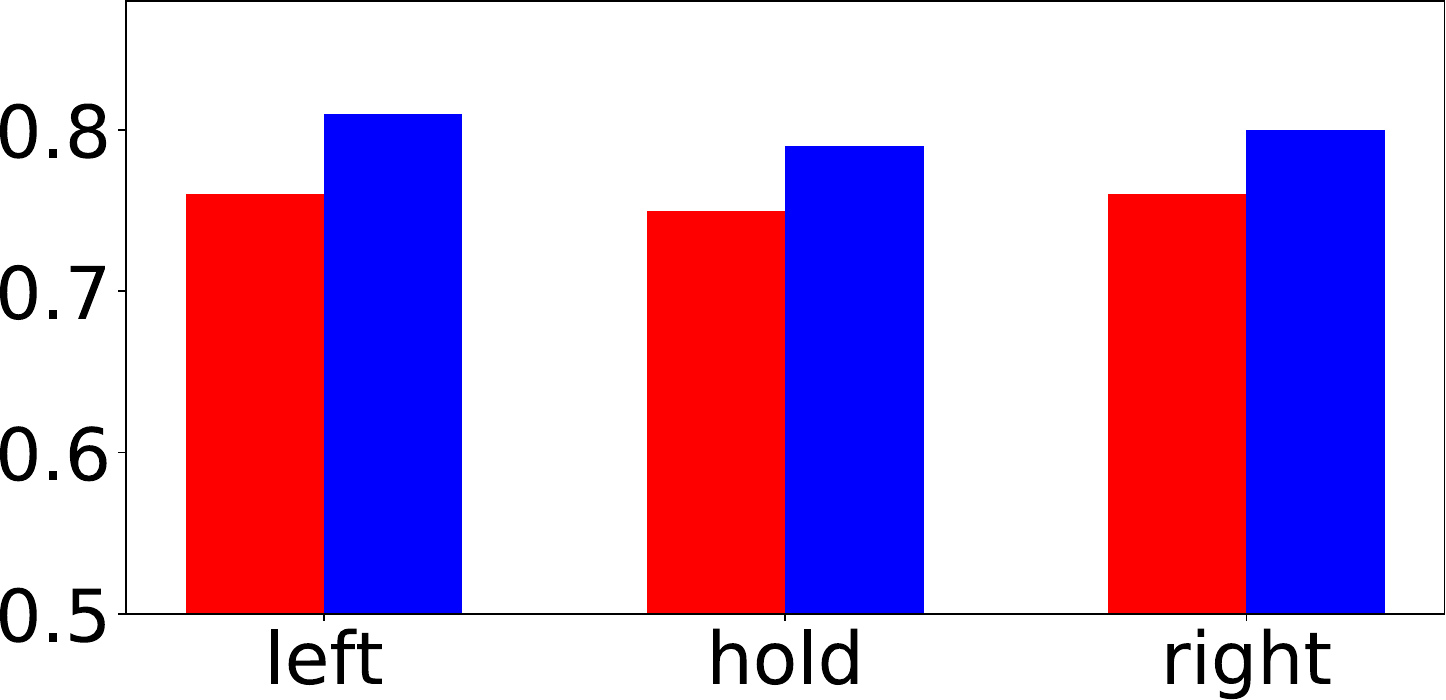}
  \includegraphics[width=0.22\textwidth]{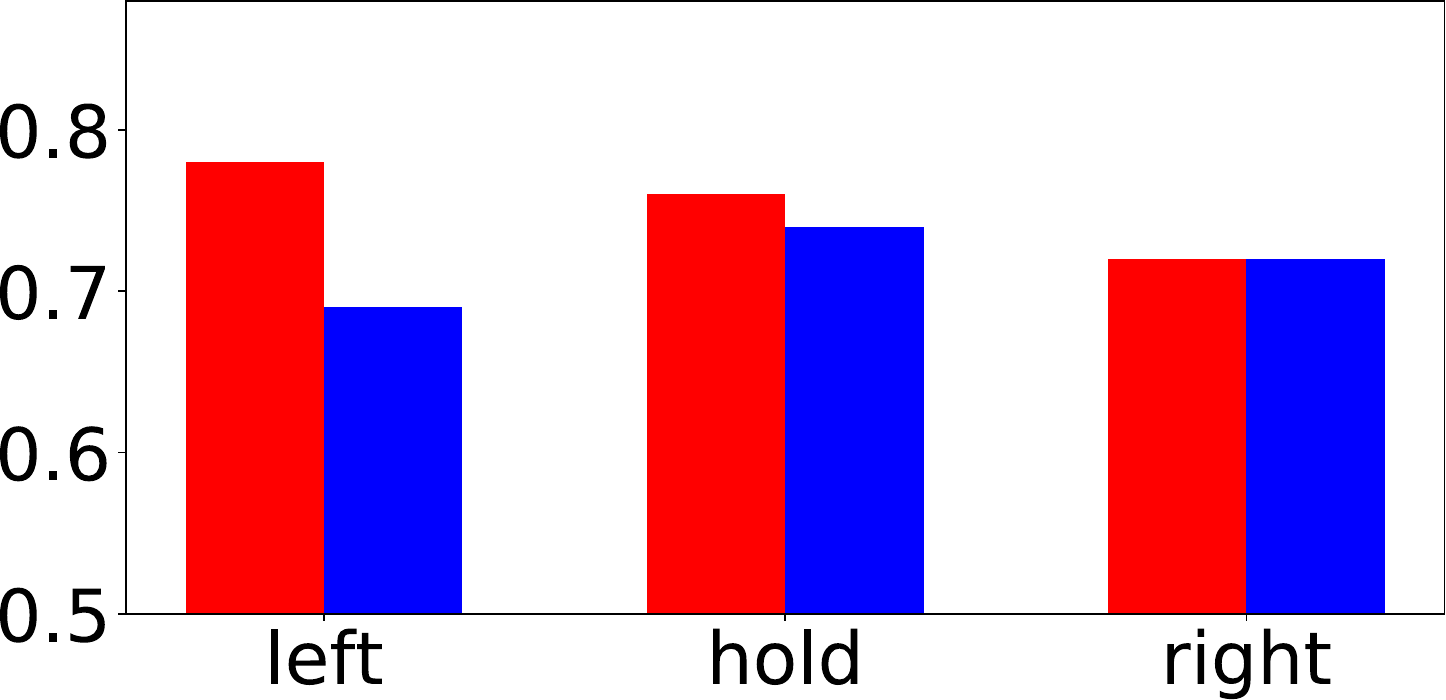} 
  \includegraphics[width=0.22\textwidth]{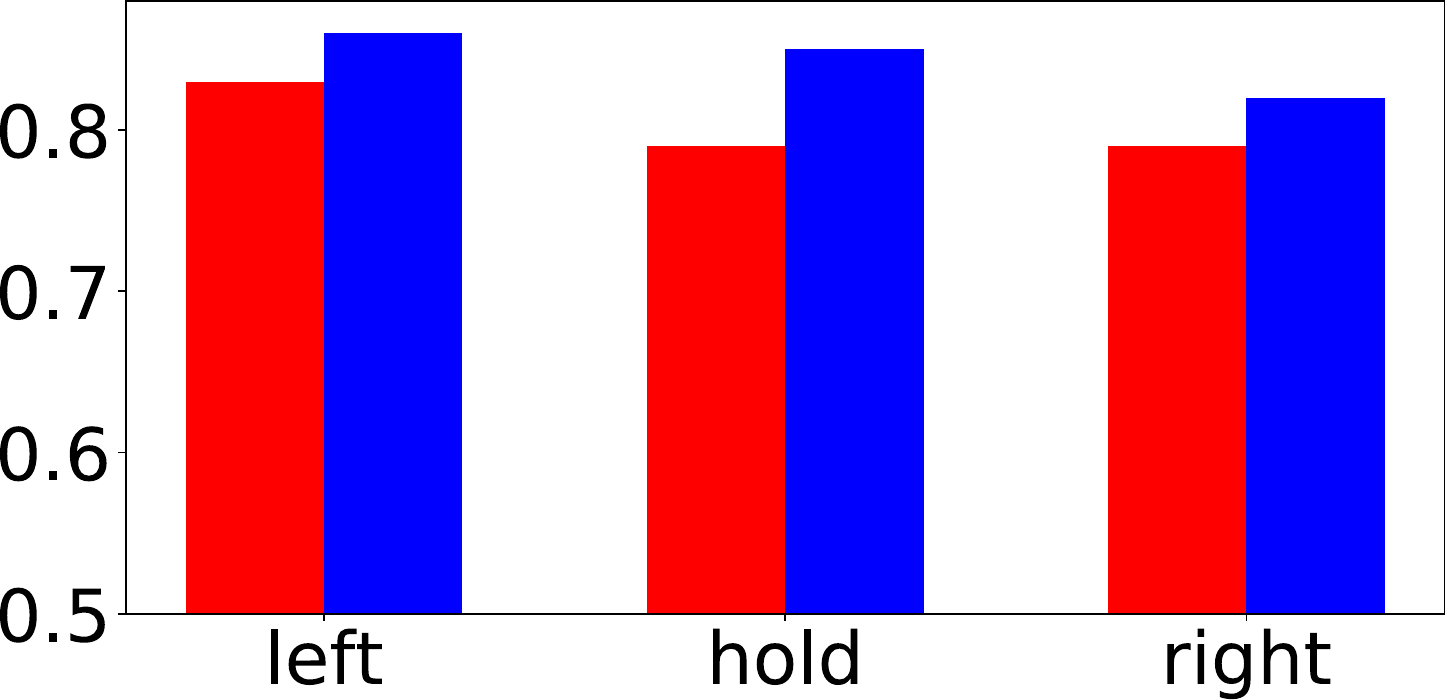}
  \includegraphics[width=0.22\textwidth]{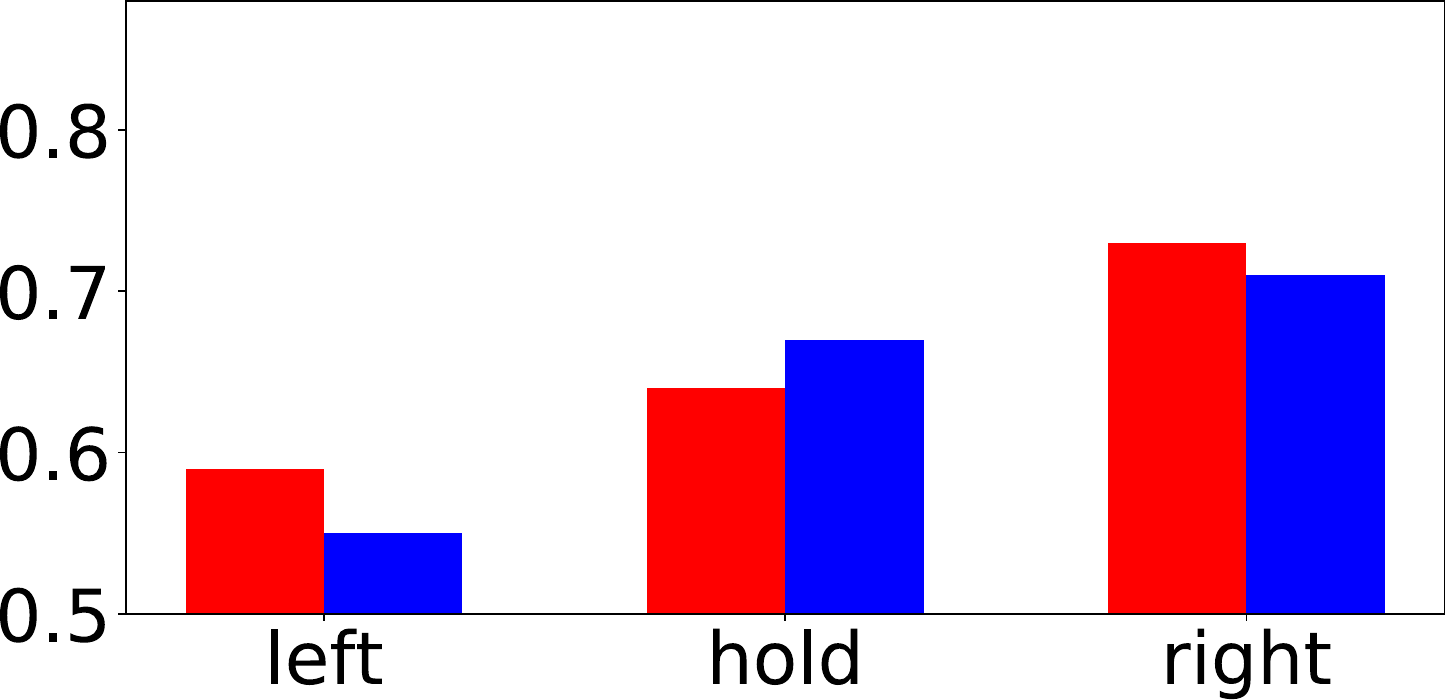}
  \vspace{-0.1cm}
  \caption{Results generated by the NPST algorithm with ``nervous'' \emph{style} (top) and ``fall'' \emph{style} (bottom). Red bars represent the outputs given by the original \emph{content} network. Blue bars represent the output given by each of the networks obtained with the NPST algorithm. The game screenshots are the results of executing the NPST generated policy.}
  \label{fig:catch-style}
\end{figure*}

Fig. \ref{fig:catch-style} depicts intermediate time steps defined by the NPST generated policy. The screenshots, in the first and third row of the figure, depict the state of the game corresponding to some randomly chosen time steps. These game states were generated following the generated policy. The same time steps were chosen for both \emph{styles}. The bar graphs compare the Q-values obtained using the Vanilla Content DQN and the NPST Generated DQN for each of the intermediate states and styles.

\section{``Grid-world paint'' experiment}
\label{experiments2}

The second set of experiments consists in a ``Grid-world paint'' scenario.
The physical agent is the humanoid robot TEO, as shown in Fig. \ref{fig:painting}, which can move its end-effector vertically and horizontally in the Cartesian space.
A large monitor emulates painting by changing large pixel colors upon presence of the robot end-effector.
These pixels form a 16x16 grid.
The goal is to reach a target pixel defined in the vertical center at the right side of the monitor.

\subsection{Experimental setup}

Three different sets of five expert demonstrations are performed. 
These sets correspond to three different actions: 
one \emph{content} action, and two different \emph{style} actions (``nervous'' and ``fall''). The \emph{content} action aims to win the game by fulfilling the goal of reaching the target pixel.
The first \emph{style} imitates a ``nervous'' behaviour or mood, tending to perform moves of one pixel around the same position.
The second \emph{style} imitates a ``fall'' movement, always tending towards the bottom side of the monitor.

The Maximum Entropy Deep IRL algorithm presented in section \ref{BG} is again used to obtain the reward functions that define these actions. These reward functions are then used to train the networks that are used with the NPST algorithm. For the IRL algorithm, a hand-crafted feature vector $\phi(s)$ and latent state space $s$ are selected. In the case of the \emph{content} demonstrations, $s$ is a 256 element vector resulting from performing a flatten operation over the 16x16 pixel grid, and $\phi(s)$ classifies if the current pixel is the target pixel. 
The same latent state space $s$ is used with the ``fall'' \emph{style} demonstrations. 
Here, $\phi(s)$ classifies if the pixel belongs to the bottom row of the grid. In the case of the ``nervous'' \emph{style} demonstrations, the latent state space $s$ is a $16^3$ element vector, where 16 is the number of possible vertical positions of the pixels, and 3 is the last three time steps. 
The feature vector $\phi(s)$ classifies if a movement was performed with the same starting and ending pixel.

For each of the three sets of demonstrations, 5 iterations of the IRL algorithm are executed.
An IRL discount factor ($\gamma$) of 0.9 is used with a learning rate of 0.01. Three reward functions $R(s)$ are obtained corresponding to the three base actions proposed (\emph{content}, ``nervous'' and ``fall''). For each of these base actions, three different neural networks, corresponding to the three architectures proposed, are trained using the same $R(s)$. These networks are trained using Q-learning and referred in the experiments as the Vanilla neural networks. These Vanilla neural networks are the base neural networks that will later be used to define the \emph{content} and \emph{style} in the NPST step. Experimental results obtained with these Vanilla networks are added as a baseline.  The same Vanilla Content neural networks are introduced for the transferring of both of the \emph{styles}. The input of the networks is the raw 16x16 pixel grid of the monitor, and the outputs correspond to the Q-values assigned to the three possible actions (stay still, go left, and go right). 

\begin{table}[h!]
\centering
\caption{Hyperparameters for the ``Grid-world paint'' scenario.}
\label{tab:grid-world}
\begin{tabular}{ll}
%\hline
Hyperparameter                 & Setting                             \\ \hline
\emph{Shared between architectures}                          &                                     \\
\hspace{3ex}Activation                                                        & ReLU \cite{nair2010}   \\ 
\hspace{3ex}Initialization                                                    & Normal distribution  \\
\emph{DQN architecture}                                                       &                                     \\
\hspace{3ex}Input shape                                                       & (16, 16, 4) \\
\hspace{3ex}Layers                                                            & (CL, CL, CL, FC, FC)   \\
\hspace{3ex}Layers configuration (Size, Kernel, Strides)                      & ((32,8,4),(64,4,2),(64,3,1),(512,-,-),(4,-,-))    \\
%\hspace{3ex}Kernel                            & [8,4,3]    \\
%\hspace{3ex}Strides                           & [4,2,1]    \\

\emph{SQN architecture}                                                        &                                     \\
\hspace{3ex}Input shape                                                        & (16, 16, 4) \\
\hspace{3ex}Layers                                                             & (CL, FC, FC)   \\
\hspace{3ex}Layers configuration (Size, Kernel, Strides)                       & ((32,8,4),(512,-,-),(4,-,-))    \\
%\hspace{3ex}Kernel                            & [8]    \\
%\hspace{3ex}Strides                           & [4]    \\

\emph{DRQN architecture}                                                       &                                     \\
\hspace{3ex}Input shape                                                        & (16, 16, 1) \\
\hspace{3ex}Layers                                                             & (CL, CL, CL, LSTM, FC)   \\
\hspace{3ex}Layers configuration (Size, Kernel, Strides)                       & ((32,8,4),(64,4,2),(64,3,1),(256,-,-),(4,-,-))    \\
%\hspace{3ex}Kernel                            & [8,4,3]    \\
%\hspace{3ex}Strides                           & [4,2,1]    \\

\emph{Q-learning and NPST}                         &                                     \\
\hspace{3ex} Grid-map size                     & 16x16                               \\
\hspace{3ex} Number of input time steps            & 4 (1 for DRQN)                                   \\
\hspace{3ex} Optimizer                      & Adam \cite{Kingma2014}    \\
\hspace{3ex} Loss function                  & Mean Squared Error                  \\
\hspace{3ex} Number of actions              & 4                                   \\
\hspace{3ex} Discount ($\gamma$)                   & 0.99                                \\
\hspace{3ex} Experience Replay size         & 50000                                \\
\emph{Q-learning}                  &                                     \\
\hspace{3ex} Learning Rate                & 1e-6                                 \\
\hspace{3ex} Initial Epsilon                & 0.9                                 \\
\hspace{3ex} Final Epsilon                  & 0.01                              \\
\hspace{3ex} Epsilon gradient               & Lineal                              \\
\hspace{3ex} Batch size                     & 32                                  \\
\hspace{3ex} Exploration Epochs             & 100                                 \\
\hspace{3ex} Training Epochs                & 5000                                \\
\emph{NPST Algorithm}                         &                                     \\
\hspace{3ex} Learning Rate                  & 0.01                                 \\
\hspace{3ex} Number of iterations (N)       & One full Grid-world episode \\
\hspace{3ex} Batch size                     & 100                                 \\
\hspace{3ex} L-BFGS-B internal iterations & 1                                
\end{tabular}%
\end{table}

The hyperparameters used for training the neural networks and performing the NPST algorithm are depicted in Table \ref{tab:grid-world}. The results for the generated NPST actions depicted in the following section are the average of 10 repetitions of the NPST algorithm.

\begin{table}[h!]
\centering
\caption{Experimental results for the ``Grid-world paint'' action introducing the ``nervous'' \emph{style}.}
\label{gridworld_nerv}
\begin{tabularx}{1\textwidth}{|c *{7}{|Y}|}
	 \hline
	 Actions & $\mathcal{L}_{content}$ & $\mathcal{L}_{style}$  & $Nervous$ $Moves$ & $Average$ $Steps$ & $Wins(\%)$ & $Partial$ $Wins(\%)$ \\ \hline
     Vanilla Content DQN  & --- & 1064.41 & 0 & 15 & 100 & 100 \\ \hline
     Vanilla Content SQN  & --- & 1370.71 & 0 & 37 & 50 & 100 \\ \hline
	 Vanilla Content DRQN  & --- & 2205.27 & 0 & 15 & 100 & 100 \\ \hline
	 Vanilla Nervous Style DQN  & 20.41 & --- & 580 & 60 & 0 & 0 \\ \hline
	 Vanilla Nervous Style SQN  & 31.34 & --- & 299 & 60 & 0 & 0 \\ \hline
	 Vanilla Nervous Style DRQN  & 22.67 & --- & 599 & 60 & 0 & 0 \\ \hline
	 NPST Nervous Generated DQN   & 127.08 & 3.20 & 46 & 58 & 20 & 80 \\ \hline
	 NPST Nervous Generated SQN   & 14.50 & 3.75 & 28 & 60 & 0 & 60 \\ \hline
	 NPST Nervous Generated DRQN   & 1.66 & 5.16 & 19 & 55 & 20 & 100 \\ \hline
	 
	\end{tabularx}

\end{table}

\begin{table*}[h!]
\centering
\caption{Experimental results for the ``Grid-world paint'' action introducing the ``fall'' \emph{style}.}
\label{gridworld_fall}
\begin{tabularx}{1\textwidth}{|c *{7}{|Y}|}
	 \hline
	 Actions & $\mathcal{L}_{content}$ & $\mathcal{L}_{style}$  & $Nervous$ $Moves$ & $Average$ $Steps$ & $Wins(\%)$ & $Partial$ $Wins(\%)$ \\ \hline
     Vanilla Content DQN  & --- & 1068.48 & 0 & 15 & 100 & 100 \\ \hline
     Vanilla Content SQN  & --- & 1380.58 & 0 & 37 & 50 & 100 \\ \hline
	 Vanilla Content DRQN  & --- & 2209.30 & 0 & 15 & 100 & 100 \\ \hline
	 Vanilla Fall Style DQN  & 115.50 & --- & 0 & 60 & 0 & 0 \\ \hline
	 Vanilla Fall Style SQN  & 78.63 & --- & 0 & 60 & 0 & 0 \\ \hline
	 Vanilla Fall Style DRQN  & 72.28 & --- & 0 & 60 & 0 & 0 \\ \hline
	 NPST Fall Generated DQN   & 14.03 & 3.05 & 43 & 60 & 0 & 40 \\ \hline
	 NPST Fall Generated SQN   & 10.14 & 3.47 & 46 & 58 & 20 & 50 \\ \hline
	 NPST Fall Generated DRQN   & 2.60 & 4.43 & 10 & 51 & 40 & 100 \\ \hline
	\end{tabularx}
\end{table*}

\subsection{Results}

The results obtained with the NPST algorithm are depicted in Table~\ref{gridworld_nerv} for the case of transferring the ``nervous'' \emph{style} and Table~\ref{gridworld_fall} for the case of transferring the ``fall'' \emph{style}. 
In these experiments, the $Nervous$ $Moves$ parameter measures the number of times the agent performed a full vertical direction swap (i.e. going upwards then downwards then upwards again). The $Average$ $Steps$ parameter measures the average number of steps performed by the agent per episode. Finally, the $Partial$ $Wins(\%)$ parameter measures the number of times the agent ended the episode in the same column as the target. This parameter was introduced to complement the $Wins(\%)$ parameter due to the low tolerance of error defined by the environment. 

The number of Nervous Moves performed was increased when introducing the ``nervous'' \emph{style} in the DQN and DRQN architectures with respect the Vanilla Content and Vanilla Fall networks. One hypothesis behind the high number of $Nervous$ $Moves$ obtained in the ``fall'' networks is due to the combination of the ``fall'' \emph{style}  and \emph{content}. The ``fall'' \emph{style} tries to constantly reach the bottom while the \emph{content} tries to bring up the agent to the middle position where the target is located. This provokes multiple vertical direction changes. Similar to what happened in the ``Catch-ball'' game scenario, the number of wins decreased when introducing the \emph{styles} with respect the Vanilla Content networks but increased with respect the Vanilla Style networks.  The $\mathcal{L}_{content}$ obtained with the NPST algorithm using the DQN architecture was unusually high due to some outliers produced in the first steps of the NPST execution with unusually high $\mathcal{L}_{content}$ values. As in the ``Catch-ball'' scenario, the results obtained with the generated control policy are a combination of the \emph{content} and \emph{style} policies.

\begin{figure*}[h!]
  \centering
  \includegraphics[width=0.7\textwidth]{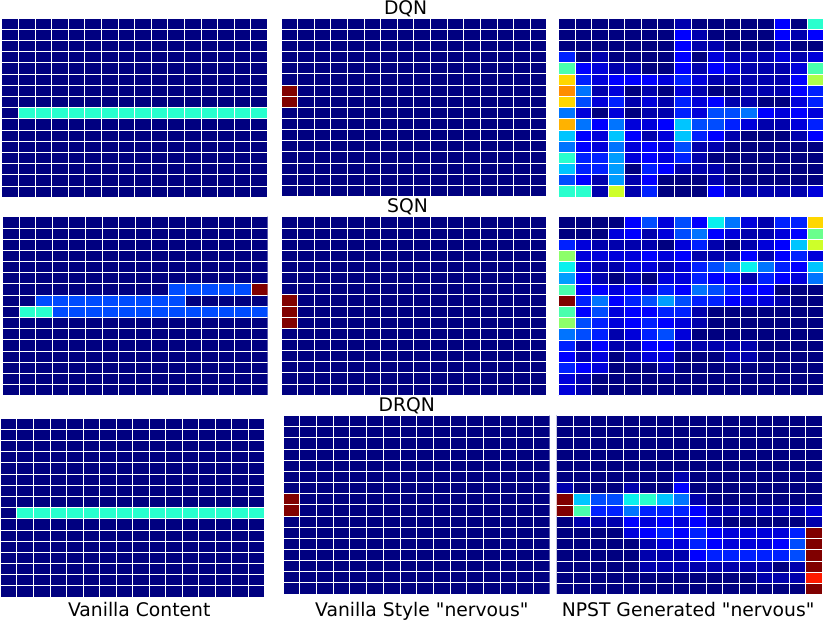}
  \caption{Robot end-effector heatmap results for transferring the ``nervous'' \emph{style}. Each row corresponds to a different network architecture. Each column corresponds to a different action. Warm colors represent monitor pixels that are recurrently visited. Cold colors represent the less visited pixels. The graphs show the cumulative results over the 10 repetitions performed. The color scale of the heatmap goes from 0 to 50. Pixels with values higher than 50 are capped to this value}
  \label{fig:heatmap-nerv}
\end{figure*}

\begin{figure*}[h!]
  \centering
  \includegraphics[width=0.7\textwidth]{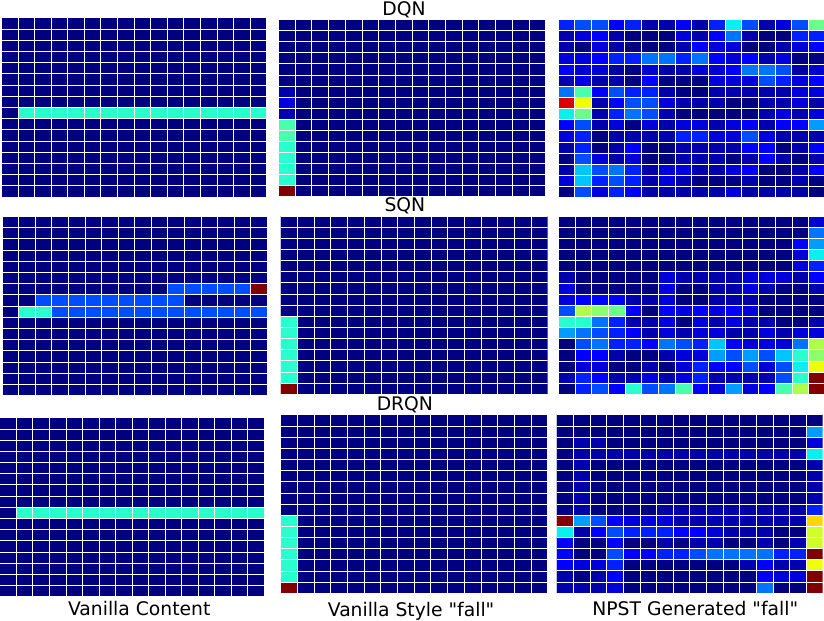}
  \caption{Robot end-effector heatmap results for transferring the ``fall'' \emph{style}. Each row corresponds to a different network architecture. Each column corresponds to a different action. Warm colors represent monitor pixels that are recurrently visited. Cold colors represent the less visited pixels. The graphs show the cumulative results over the 10 repetitions performed. The color scale of the heatmap goes from 0 to 50. Pixels with values higher than 50 are capped to this value.}
  \label{fig:heatmap-fall}
\end{figure*}

Fig. \ref{fig:heatmap-nerv} and Fig. \ref{fig:heatmap-fall}  are heatmaps representing the most visited states for each of the generated policies. Warmest colors depict the most recurrently visited pixels, and the coldest colours represent the least visited pixels. As expected, in these figures, the policies transferred with the ``fall'' \emph{style} show a clear preference for the bottom states in the case of the DQN and DRQN architecture. The same preference is also depicted for the DRQN architecture while transferring the ``nervous'' \emph{style}. The rest of the policies do not depict any relevant preference in terms of vertical position.

\section{Discussion}
\label{sect:Discussion}

One of the first ideas that has to be taken in account when studying the results proposed in this paper is that by introducing Style Transfer a new optimization problem is defined. With Style Transfer, the goal is not to find the optimal policy for the execution of the content action, but to find the optimal policy that is able to execute the content action with the selected \emph{style}. This is relevant in the case of the results obtained in the experiments proposed. Here, the percentage of wins was reduced with the introduction of the two proposed styles. This reduction, however, came with a decrease in the total Style Transfer loss depicting a more optimal policy for the Style Transfer problem. The reduction in the percentage of wins can be therefore depicted as an expected or even desired behavior. Players with some emotional bias are expected to have a lower performance when executing the action that players without it. 
%This is the case for example of a nervous player where the expected behavior is not to win all the games.

The qualitative and quantitative results obtained in this paper show how NPST is able to transfer and combine the behavior of different policies into new ones. This allows the introduction of future applications introducing \emph{styles} in robotic actions. For social robot applications, these \emph{styles} can be defined as different human emotions. More pragmatic applications may define these \emph{styles} as a way to improve the performance of the robot using only a set of demonstrated expert trajectories that can be generalized to the full range of robotic actions.

The formulation of the NPST algorithm presents the limitation of requiring to perform a new execution each time a new trajectory has to be generated. The generated policy is defined by the output of the NPST execution. At the same time, the NPST algorithm presents some relevant advantages with respect other State of the Art methods. One critical advantage is that the NPST algorithm works without requiring an additional loss network. The same network is used to generate the trajectory and perform the loss extraction. This reduces computational times and the overall complexity of the Style Transfer process. In addition to this, the NPST framework is designed to work with robots. Within the authors knowledge, this is the first time Style Transfer is introduced within a Reinforcement Learning framework for the generation of \emph{style} control policies and its execution with robotic actions. 

\section{Conclusions}
\label{conclusions}

Style Transfer aims to transform a certain input, adapting it via a certain \emph{style} without changing the original \emph{content}.
It has been proposed and succesfully introduced in a number of fields (fine arts, natural language processing, and fixed trajectories).
By means of this work, a Neural Policy Style Transfer (NPST) algorithm has been proposed to perform Style Transfer with control policies.

The control policies are defined by neural networks that express Reinforcement Learning Q-value functions. 
The generated action is initialized to have the same model and number of parameters as the \emph{style} action. The input must correspond to the observation or state space, and the output must express the Q-value function. The base neural networks are trained via Maximum Entropy Deep IRL algorithms that learn reward functions, taught by human demonstrators and NPST-ready.

Two sets of experiments were performed in this paper, each corresponding to the two scenarios presented.
The ``Catch-ball'' game is inspired by the Deep Reinforcement Learning classical Atari games;
and the ``Grid-world paint'' scenario includes a full-sized humanoid robot, equivalent to a grid-world problem in the real world, based on previous works of the authors.
In both sets of experiments, the results show a clear influence in the execution of the policies after transferring each of the \emph{styles}. Three different architectures were introduced to test the NPST algorithm. 
%From these three architectures, the DQN architecture achieves the best overall performance in the proposed experiments. 
The results show a clear influence of the transferred \emph{style} in the generated action while keeping the \emph{content} goal. The resulting control policy introduces elements of both the proposed \emph{style} and the defined \emph{content}.

\section{Acknowledgment}
The research leading to these results has received funding from RoboCity2030-DIH-CM Madrid Robotics Digital Innovation Hub (“Robótica aplicada a la mejora de la calidad de vida de los ciudadanos. fase IV”; S2018/NMT-4331), funded by “Programas de Actividades I+D en la Comunidad de Madrid” and cofunded by Structural Funds of the EU. The authors thank Bartek \L{}ukawski for his valuable collaboration.

%\section*{References}
%APA STYLE
\bibliography{./st2020.bib}
\bibliographystyle{apalike}

\end{document}